\documentclass{article}

\usepackage[preprint]{neurips_2023}
\usepackage[utf8]{inputenc} 
\usepackage[T1]{fontenc}    
\usepackage[dvipsnames]{xcolor}
\usepackage[colorlinks=true,linkcolor=Purple,citecolor=Emerald]{hyperref}
\usepackage{url}            
\usepackage{booktabs}       
\usepackage{amsfonts}       
\usepackage{nicefrac}       

\usepackage{microtype}      
\usepackage{caption}
\usepackage{multirow}
\usepackage{wrapfig}
\usepackage{xspace}
\usepackage{tcolorbox}
\usepackage{colortbl}
\usepackage{graphicx}
\usepackage{fancyhdr,lipsum}
\usepackage[export]{adjustbox}
\usepackage{enumitem} 

\tcbuselibrary{most}

\title{Knowledge Fusion of Chat LLMs: A Preliminary Technical Report}

\author{Fanqi Wan, Ziyi Yang, Longguang Zhong, Xiaojun Quan\thanks{$\;\;$Corresponding authors.}, Xinting Huang, Wei Bi \\\\
\text{FuseAI Research Team}
\\
}

\begin{document}
\maketitle


\begin{abstract}
Recently, \textsc{FuseLLM} introduced the concept of knowledge fusion to transfer the collective knowledge of multiple structurally varied LLMs into a target LLM through lightweight continual training. In this report, we extend the scalability and flexibility of the \textsc{FuseLLM} framework to realize the fusion of chat LLMs, resulting in \textsc{FusionChat}. \textsc{FusionChat} comprises two main stages. Firstly, we undertake knowledge fusion for structurally and scale-varied source LLMs to derive multiple target LLMs of identical structure and size via lightweight fine-tuning. Then, these target LLMs are merged within the parameter space, wherein we propose a novel method for determining the merging weights based on the variation ratio of parameter matrices before and after fine-tuning. We validate our approach using three prominent chat LLMs with diverse architectures and scales, namely \texttt{NH2-Mixtral-8x7B}, \texttt{NH2-Solar-10.7B}, and \texttt{OpenChat-3.5-7B}. Experimental results spanning various chat domains demonstrate the superiority of \texttt{\textsc{FusionChat}-7B} across a broad spectrum of chat LLMs at 7B and 34B scales, even surpassing \texttt{GPT-3.5 (March)} and approaching \texttt{Mixtral-8x7B-Instruct}. 

\end{abstract}

\section{Introduction}

Large language models (LLMs) such as GPT~\citep{brown2020language} and LLaMA~\citep{touvron2023llama-2} series have demonstrated remarkable success across a wide range of natural language processing (NLP) tasks. 
However, the computational resources and time costs associated with LLM development remain prohibitively high for most entities. 
Despite the structural and functional differences among LLMs, they often exhibit similar capabilities across various tasks. Therefore, moving beyond the traditional approach of training a LLM from scratch, an alternative option is to combine existing LLMs into a new, more powerful one, which is termed \emph{knowledge fusion of LLMs} by ~\cite{wan2024knowledge}. 
If successful, this fusion not only reduces the initial training costs but also enables the combined model to leverage the strengths of multiple LLMs.

The endeavor to integrate the capabilities of multiple models has been a long-standing pursuit. 
For example, ensemble methods~\citep{littlestone1994weighted, jiang2023llm} directly aggregate the outputs of different models to enhance prediction performance and robustness. 
However, this approach requires maintaining multiple trained models and executing each during inference, which is inefficient for LLMs due to their substantial memory and inference time requirements. 
Another approach is to directly merge several neural networks into a single network through arithmetic operations in the parameter space~\citep{gupta2020stochastic}. This approach typically assumes uniform network architectures and seeks to merge the parameters of different neural networks either through manual merging weights~\citep{wortsman2022model, yadav2023ties} or by automatically obtaining merging weights based on model gradients or representations of additional data~\citep{matena2022merging, jin2022dataless}. Recently, \textsc{FuseLLM}~\citep{wan2024knowledge} introduced a new paradigm for integrating the capabilities of multiple LLMs. This approach externalizes the knowledge of multiple source LLMs using their generated probability distribution matrices and transfers their collective knowledge into a target LLM through lightweight continual training. Consequently, \textsc{FuseLLM} facilitates the fusion of multiple pre-trained LLMs with distinct architectures into a unified LLM.

In this study, we extend the framework of \textsc{FuseLLM} to fuse multiple chat LLMs with diverse architectures and scales, leading to the development of \textsc{FusionChat}, which comprises two main stages. Firstly, it conducts knowledge fusion for source LLMs with varying structures and scales to derive multiple target LLMs of identical structure and size. To this end, \textsc{FusionChat} follows the idea of \textsc{FuseLLM} but adopts a pairwise knowledge fusion strategy. Secondly, these target LLMs are merged within the parameter space to incorporate the collective knowledge and respective advantages from source LLMs. For merging, we introduce \textsc{VaRM} (\textbf{Va}riation \textbf{R}atio \textbf{M}erge), a novel method for determining the merging weights based on the variation ratio of parameter matrices before and after fine-tuning. In contrast to previous approaches, \textsc{VaRM} enables the automatic allocation of distinct weights to each parameter matrix based on the variation ratio of updates during fine-tuning. This facilitates merging LLMs with fine-grained weights without requiring additional training efforts.

\textsc{FusionChat} offers superior scalability compared to \textsc{FuseLLM}. Firstly, while \textsc{FuseLLM} limits its exploration to LLMs of the same size as the target LLM, \textsc{FusionChat} delves into the fusion of source chat LLMs with varying sizes. This broader scope allows for greater adaptability to diverse model configurations and requirements. Secondly, the framework of \textsc{FuseLLM} does not seamlessly support the inclusion of new source LLMs as it requires the combination of distribution matrices from all source LLMs during continual training. In contrast, integrating a new source LLM at any scale in \textsc{FusionChat} is plug-and-play, requiring only obtaining a target LLM from the new source LLM and merging it with the existing version of \textsc{FusionChat}. Given the frequent updates of chat LLMs in the open-source community\footnote{There are 7300+ chat LLMs available on HuggingFace as of drafting this report.}, \textsc{FusionChat} appears to be more promising for the fusion of chat models.

To empirically demonstrate the effectiveness of \textsc{FusionChat}, we implement \textsc{FusionChat} using three representative open-source chat LLMs for fusion: \texttt{NH2-Mixtral-8x7B}~\citep{jiang2024mixtral}, \texttt{NH2-Solar-10.7B}~\citep{kim2023solar}, and \texttt{OpenChat-3.5-7B}~\citep{wang2023openchat}. Experimental results on MT-Bench~\citep{zheng2023judging}, a cutting-edge benchmark consisting of eight different domains to assess chat LLMs' multi-turn dialogue ability, confirm that \textsc{FusionChat} outperforms all the source LLMs and fine-tuned baselines at 7B and 10.7B scales, even approaching the 8x7B MoE source LLM. Moreover, among all the merging methods, the proposed \textsc{VaRM} achieves the best performance, indicating the efficacy of merging weights based on the variation ratio of updates.

\section{Related Work}
\label{sec:related_work}
\paragraph{Model Fusion} The fusion of capabilities from diverse models has been a long-standing objective, with existing approaches mainly falling into three categories. 
Firstly, the traditional technique of model \emph{ensemble} combines the outputs of multiple models to enhance overall system performance~\citep{littlestone1994weighted,sagi2018ensemble}. 
Note that this technique doesn't involve the explicit merging of multiple models into a new one. 
Common methods for model ensemble typically employ weighted averaging~\citep{littlestone1994weighted} or majority voting~\citep{monteith2011turning}  to consolidate predictions from various models. 
Recently, \cite{jiang2023llm} introduced an ensemble framework designed to leverage the diverse strengths of multiple open-source LLMs. 
This framework first employs a pairwise comparison method to detect subtle distinctions among candidate outputs. 
Then, it combines the top-ranked candidates to produce an enhanced output.

Secondly, \emph{model merging} presents another approach that facilitates model fusion within the parameter space. 
\cite{wortsman2022model} combined multiple models, obtained through different strategies or configurations, through a linear weighted average of parameters, resulting in enhanced overall performance.
Likewise, \cite{shoemake1985animating} and \cite{ilharco2022editing} integrated the capabilities of distinct models by employing spherical linear interpolation and task arithmetic to merge model parameters.
To avoid redundant parameter interference, \cite{yadav2023ties} and \cite{yu2023language} suggested pruning low-amplitude varying parameter values before model merging.
Furthermore, \cite{matena2022merging} and \cite{jin2022dataless} incorporated supplementary data to compute merging weights based on model gradients or representations, eliminating the need for hyperparameter tuning.

Lastly, \textsc{FuseLLM}~\citep{wan2024knowledge} presents a new paradigm for knowledge fusion of multiple LLMs, which leverages the probabilities distribution matrices generated from source LLMs to transfer the collective knowledge and respective advantages into a target LLM. 
In comparison to the model ensemble method, which requires the parallel deployment of multiple models, and the model merging approach, which is generally limited to models with identical architectures, \textsc{FuseLLM} supports the fusion of multiple source LLMs with different architectures into a target LLM.

\paragraph{Knowledge Distillation} Knowledge distillation \citep{hinton2015distilling}, initially proposed for model compression, involves training a student model under the guidance of one or more teacher models. In the NLP community, knowledge distillation has been widely applied to text classification tasks. These applications include training the student model to replicate the teacher's output distribution~\citep{sanh2019distilbert, turc2019well}, as well as features~\citep{sun2019patient, jiao2020tinybert} and relations~\citep{wang2020minilm} derived from intermediate layers of the teacher model. In the realm of text generation, the conventional approach focuses on minimizing the KL divergence between the student and teacher generation distributions. 
This is achieved by using the teacher's probability distributions at each time step as supervision~\citep{khanuja2021mergedistill,gu2023knowledge, agarwal2023gkd} or by directly training on the teacher's generated texts~\citep{peng2023instruction, xu2023wizardlm}.

\section{Knowledge Fusion of Chat Models}

The core concept of \textsc{FusionChat} comprises two stages. Firstly, it externalizes and transfers the knowledge and capabilities inherent in source chat LLMs to multiple target LLMs of the same structure and size. Secondly, these target LLMs are incorporated into a final fused LLM through model merging. 

Specifically, considering $K$ source chat LLMs \( \{ \mathcal{M}^s_i\}_{i=1}^{K} \) with varying architectures and scales, \textsc{FusionChat} first specifies a source LLM $\mathcal{M}^s_v$ as the \emph{pivot} and then applies pairwise knowledge fusion for the pivot and each of the rest LLMs, obtaining ($K-1$) target LLMs \( \{ \mathcal{M}^t_j\}_{j=1}^{K-1} \) which share the same architecture and initial parameters as the pivot LLM. To perform the pairwise knowledge fusion, \textsc{FusionChat} prompts these source LLMs using a compact and representative training dataset $\mathcal{D}$ to showcase their inherent knowledge by predicting the next token. The resulting probabilistic distribution matrices are then utilized to perform pairwise knowledge fusion through lightweight fine-tuning as \textsc{FuseLLM}~\citep{wan2024knowledge}. After that, the ($K-1$) target LLMs are combined in the parameter space using a specific merging method to yield the fused LLM \(\mathcal{M}^f\). To incorporate fine-grained advantages of source LLMs, we introduce \textsc{VaRM} (\textbf{Va}riation \textbf{R}atio \textbf{M}erge) to determine the merging weights based on the variation ratio of parameter matrices before and after fine-tuning. In the following sections, we provide a brief introduction to the preliminaries, followed by a detailed description of the pairwise knowledge fusion and model merging in \textsc{FusionChat}.


\subsection{Preliminaries}

Let us consider a text sequence $q$ of length $N$, which is sampled from the training dataset $\mathcal{D}$.
The sequence preceding the $i$th token is represented by $t_{<i} = (t_{1}, t_{2}, \ldots, t_{i-1})$.
The causal language modeling (CLM) objective for training a language model parameterized by $\theta$ is defined as minimizing the negative log-likelihood:
\begin{equation}\label{eqn:token_level_clm}
\mathcal{L}_{\text{CLM}} = -\mathbb{E}_{q \sim \mathcal{D}} \left[ \sum_{i} \log p_{\theta}(t_i | t_{<i}) \right],
\end{equation}

where $p_{\theta}(t_i | t_{<i})$ is the model's predicted probability for the $i$th token given the preceding tokens.

To facilitate the fine-tuning of chat LLMs, wherein the text sequence $q$ often consists of a multiple-turn dialogue between a user and an assistant, we follow previous works~\citep{vicuna2023, wan2023explore} and mask the user instructions when calculating the training loss $\mathcal{L}_{\text{CLM}}$.

The above objective decomposes sequence likelihood into token-level cross-entropy losses, comparing each token's predicted distribution to its one-hot representation. To provide a more generalized perspective, we reframe this token-level view into a sequential distribution format. 
Specifically, for the text sequence $q$, we aggregate token-level predictions to form a probabilistic distribution matrix, $\mathbf{P}_q^\theta\in \mathbb{R}^{N \times V}$, where the $i$-th row represents the distribution predicted by the model for the $i$th token over the vocabulary of size $V$. The CLM objective can then be interpreted as reducing the discrepancy between $\mathbf{P}_q^\theta$ and the one-hot label matrix, $\mathbf{O}_q \in \{0, 1\}^{N \times V}$, where each row is a one-hot representation of the corresponding gold token.
Formally, the CLM objective is transformed into the following representation:
\begin{equation}\label{eqn:sequence_level_clm}
\mathcal{L}_{\text{CLM}} = -\mathbb{E}_{q \sim \mathcal{D}}\left[ \mathbb{D}(\mathbf{P}_q^\theta, \mathbf{O}_q) \right],
\end{equation}
where $\mathbb{D}(\cdot, \cdot)$ represents the discrepancy function between two matrices, and it is equivalent to Eq. \ref{eqn:token_level_clm} when implemented using the KL divergence.

\subsection{Pairwise Knowledge Fusion}
\label{sec:knowledge_fusion}
Taking this perspective on a language model, we follow \cite{wan2024knowledge} and assume that the probabilistic distribution matrix reflects certain inherent knowledge of the language model in understanding the text. 
Consequently, different probabilistic distribution matrices for the same text, originating from various LLMs, can be used to represent the diverse knowledge embedded within these models. 
Based on this assumption, the proposed \textsc{FusionChat} externalizes the knowledge of source LLMs through probabilistic modeling and performs pairwise knowledge fusion by fine-tuning target LLMs using the generated distribution matrices of the source LLMs.

Specifically, for each text sample $q$ in the training dataset $\mathcal{D}$, we first apply the provided $K$ source LLMs to obtain a set of probabilistic distribution matrices, denoted as $\{\mathbf{P}_q^{\theta_j}\}_{j=1}^{K}$, where $\theta_j$ represents the parameters of the $j$th source LLM. Note that these source LLMs may employ different tokenizers, and token alignment is often necessary for proper mapping of probabilistic distribution matrices \citep{fu2023specializing,wan2024knowledge}.
Utilizing these matrices, we externalize the knowledge from individual models into a unified space, essentially creating unified probabilistic representations over the text.

Then, pairwise knowledge fusion is conducted between the pivot LLM and each of the rest source LLMs. To achieve this, we denote the probabilistic distribution matrix generated by the pivot LLM as $\mathbf{P}_q^{\theta_v}$ and obtain a set $\{\mathbf{P}_q^j\}_{j=1}^{K-1}$ of fused matrices as follows:
\begin{equation}\label{eqn:matrix_fusion}
\mathbf{P}_q^j = \mathbb{F}\text{usion}(\mathbf{P}_q^{\theta_v}, \mathbf{P}_q^{\theta_j})|_{v\neq j},
\end{equation}
where $\mathbb{F}\text{usion}(\cdot)$ represents the function that fuses two matrices, and the resulting matrix $\mathbf{P}_q^j$ is seen as a representation of the collective knowledge and distinctive strengths of two source LLMs. Among various fusion strategies, this work employs minimum cross-entropy (MinCE) \citep{wan2024knowledge}, which empirically performs the best in both \textsc{FuseLLM} and \textsc{FusionChat}.

After that, we enforce alignment between the prediction of each target LLM $\mathcal{M}^t_j$ and the corresponding fused representation matrices $\mathbf{P}_q^{j}$. We use $\mathbf{Q}_q^{\phi_j}$ to represent the output distribution matrix of the target LLM $\mathcal{M}^t_j$ for text $q$, and then define the fusion objective for each target LLM as follows:
\begin{equation}\label{eqn:final_fusion_objective}
\mathcal{L}_{\text{Fusion}} = -\mathbb{E}_{q \sim \mathcal{D}}\left[\mathbb{D} (\mathbf{Q}_q^{\phi_j},\mathbf{P}_q^{j})\right].
\end{equation}

The overall training objective for each target LLM consists of a weighted combination of the causal language modeling objective $ \mathcal{L}_{\text{CLM}} $ and the fusion objective $ \mathcal{L}_{\text{Fusion}}$ as follows:
\begin{equation}\label{eqn:final_objective}
\mathcal{L} = \lambda\mathcal{L}_{\text{CLM}} + (1-\lambda)\mathcal{L}_{\text{Fusion}}.
\end{equation}

\subsection{Model Merging}
\label{sec:model_merge}
Given that the fused target LLMs \( \{ \mathcal{M}^t_j\}_{j=1}^{K-1} \) share identical architecture and scale while possessing diverse advantages and capabilities learned from the source LLMs, which can be further integrated in the parameters space \citep{wortsman2022model} to obtain the final fused LLM $\mathcal{M}^f$:
\begin{equation}\label{eqn:merging}
\mathcal{M}^f = \mathbb{M}\text{erge}(\{\mathcal{M}^t_1,\mathcal{M}^t_2,...,\mathcal{M}^t_{K-1}\}),
\end{equation}
where $\mathbb{M}\text{erge}(\cdot)$ denotes the function that merges multiple target LLMs into a final LLM that combines collective knowledge and distinctive strengths of these target LLMs.

To enhance the adaptability of \textsc{FusionChat}, it is essential to maintain the simplicity of the $\mathbb{M}\text{erge}$ function. Firstly, it should be capable of automatically computing the merging weights, eliminating the need for intricate hyperparameter tuning. Secondly, the merging procedure should not require the incorporation of additional data for the calculation of model gradients or representations.

Since the parameters of the target LLMs continuously evolve to align their generated distribution matrices with the corresponding source LLMs, we propose \textbf{Va}riation \textbf{R}atio \textbf{M}erge (\textsc{VaRM}) to utilize the variation ratio of parameters before and after fine-tuning each target LLM as an indicator of knowledge updates, determining its importance in the $\mathbb{M}\text{erge}$ function:
\begin{equation}\label{eqn:varn}
W_{j,m} = \frac{\mathbb{E}_{m}\Delta\theta^{2}_{j,m}}{\sum^{K-1}_{j=1}\mathbb{E}_{m}\Delta\theta^{2}_{j,m}}.
\end{equation}

where $W_{j,m}$ represents the merging weight for the parameter unit $\theta_{j,m}$ (e.g., a matrix) in the target LLM $\mathcal{M}^t_{j}$, while $\mathbb{E}_{m}\Delta\theta^{2}_{j,m}$ denotes the average squared variation of parameters in the unit $\theta_{j,m}$.

In our preliminary explorations, we have investigated several alternative approaches to determining the weights. These include replacing the square operation with the absolute operation or using softmax. However, the results indicate that none of these alternatives outperforms the current method.

In this work, we define the parameter unit for model merging at the matrix level\footnote{We discuss the influence of different merging granularities in Section \ref{sec:analysis_of_merging_granularity}.}. This approach enables the automatic allocation of distinct merging weights to each parameter matrix, thereby facilitating the integration of fine-grained advantages from multiple target LLMs into the fused LLM.

\subsection{Discussions}
The reasons why \textsc{FusionChat} does not directly follow \textsc{FuseLLM} to fuse multiple source LLMs of different structures and scales are twofold. Firstly, directly fusing all the source LLMs proves to be difficult, as evidenced by the results of \texttt{OpenChat-3.5-7B Multi} in Table \ref{tab:main_results}. Instead, \textsc{FusionChat} adopts a fuse-then-merge strategy, wherein the fusing stage employs pairwise knowledge fusion between the pivot LLM and other source LLMs, reducing the difficulty of model fusion. Secondly, \textsc{FusionChat} offers superior scalability compared to \textsc{FuseLLM}. The framework of \textsc{FuseLLM} requires the combination of distribution matrices from all source LLMs during continual training, which does not seamlessly support the inclusion of new source LLMs. In contrast, \textsc{FusionChat} supports plug-and-play integration of a new source LLM at any scale, requiring only obtaining a target LLM by fusing the new source LLM and the pivot, and then merging it with the existing version of \textsc{FusionChat}.

Moreover, the concept of knowledge fusion adopted by both \textsc{FusionChat} and \textsc{FuseLLM} shares a fundamentally similar purpose with other related topics, such as traditional model ensemble and merging techniques, as well as the recently prominent topic of mixture of experts (MoEs), because they all aim to leverage the strengths of multiple models (experts). While model ensemble and MoEs require loading multiple models (experts) during inference, which have higher memory requirements, weight merging is limited to models with identical architectures. In contrast, knowledge fusion supports the integration of multiple LLMs with diverse architectures into a single LLM without any additional memory requirement, making it appealing in terms of both flexibility and efficiency.

\section{Experiments}

In our experiments, we consider a challenging scenario for the fusion of chat LLMs, where the source LLMs exhibit minimal similarities in architectures and scales. 
Specifically, we conduct experiments with three representative open-source chat LLMs as the source LLMs, including \texttt{NH2-Mixtral-8x7B}\footnote{\url{https://huggingface.co/NousResearch/Nous-Hermes-2-Mixtral-8x7B-DPO}}~\citep{jiang2024mixtral}, \texttt{NH2-Solar-10.7B}\footnote{\url{https://huggingface.co/NousResearch/Nous-Hermes-2-SOLAR-10.7B}}~\citep{kim2023solar}, and \texttt{OpenChat-3.5-7B}\footnote{\url{https://huggingface.co/openchat/openchat_3.5}}~\citep{wang2023openchat}. 
As for the pivot LLM, which also serves as the starting point for target LLMs, we opt for \texttt{OpenChat-3.5-7B} due to its balanced scale and performance. We then apply pairwise knowledge fusion as introduced in Section \ref{sec:knowledge_fusion} to obtain two target LLMs \texttt{OpenChat-3.5-7B Mixtral} and \texttt{OpenChat-3.5-7B Solar}. 
Finally, we merge \texttt{OpenChat-3.5-7B Mixtral} and \texttt{OpenChat-3.5-7B Solar} by our \textsc{VaRM} method (Section \ref{sec:model_merge}) to obtain the final \texttt{\textsc{FusionChat}-7B}.
To assess the performance of \textsc{FusionChat}-7B, we conduct experiments on MT-Bench\footnote{\url{https://github.com/lm-sys/FastChat/tree/main/fastchat/llm_judge}}~\citep{zheng2023judging}, a benchmark specifically designed to evaluate chat LLMs' capabilities in multi-turn dialogues across various domains.

\subsection{Experimental Setup}
\label{sec:experimental_setup}

\paragraph{Training Dataset}
To acquire the advantages of source LLMs during knowledge fusion, while mitigating catastrophic forgetting, we curated a high-quality training dataset named \textsc{FusionChat Mixture} from two sources. Firstly, 50\% of our training data is sampled from the dataset used by OpenChat\footnote{\url{https://huggingface.co/openchat/openchat_3.5\#dataset-details}}. Secondly, we collected the remaining training samples, unseen by OpenChat, from open-source communities. These two sources resulted in a collection of around 95,000 dialogues across various domains. Further details of \textsc{FusionChat Mixture} can be found in Appendix \ref{appendix: details_of_training_dataset}.

\paragraph{Training Details}
In all experiments, we train the \texttt{OpenChat-3.5-7B} using a batch size of 128 and a maximum length of 2048 on a single node with 8x40GB NVIDIA A100 GPUs for three epochs, which takes approximately 7 hours. 
The model is optimized using the AdamW~\citep{loshchilov2017decoupled} optimizer with $\beta_{1}=0.9$ and $\beta_{2}=0.999$, with gradient clipping set to 1.0 and weight decay to 0.0. 
A cosine learning rate schedule is employed, with a maximum learning rate of 5e-6 and a warmup ratio of 0.03.
We empirically set the combination weight $\lambda$ in Eq. \ref{eqn:final_objective} to 0.9. 
Our training framework is implemented based on the HuggingFace Transformers~\citep{wolf2020transformers}. 

\paragraph{Evaluation}
We evaluate \textsc{FusionChat} on MT-Bench, which comprises 80 multi-turn dialogues spanning \textit{writing}, \textit{roleplay}, \textit{reasoning}, \textit{math}, \textit{coding}, \textit{stem}, and \textit{humanities} domains. 
We adhere to the default configuration of \cite{zheng2023judging} and use GPT-4\footnote{\url{https://platform.openai.com/docs/models}} (\texttt{gpt-4-0613}) as the evaluator for the generated responses, setting the temperature to 0.0 to ensure replicability. 
The evaluation score ranges from 1 to 10, with 1 denoting the poorest quality and 10 denoting the best.
 
\paragraph{Baselines}
In our experiments, we compare our \textsc{FusionChat} with three categories of baselines. 
(i) \emph{Closed-source LLMs}: \texttt{GPT-4 (March)}, \texttt{GPT-3.5 (March)}, and \texttt{Claude-1.0}. 
(ii) \emph{Source LLMs}: \texttt{NH2-Mixtral-8x7B}, \texttt{NH2-Solar-10.7B}, and \texttt{OpenChat-3.5-7B}. 
(iii) \emph{Fine-tuned target LLMs}: \texttt{OpenChat-3.5-7B CLM}, which is fine-tuned using only the casual language modeling objective; \texttt{OpenChat-3.5-7B Multi}, which is fine-tuned using the fusion of distributions generated from multiple source LLMs~\citep{wan2024knowledge}; \texttt{OpenChat-3.5-7B Mixtral}, which is the corresponding target LLM obtained by fusing \texttt{OpenChat-3.5-7B} and \texttt{NH2-Mixtral-8x7B}; \texttt{OpenChat-3.5-7B Solar}, which is the corresponding target LLM obtained by fusing \texttt{OpenChat-3.5-7B} and \texttt{NH2-Solar-10.7B}.
We also evaluate the performance of \textsc{FusionChat} by comparing different merging methods to obtain the \emph{fused LLMs}, including \texttt{\textsc{FusionChat}-7B Linear}~\citep{wortsman2022model}, \texttt{\textsc{FusionChat}-7B SLERP}~\citep{shoemake1985animating}, \texttt{\textsc{FusionChat}-7B TA}~\citep{ilharco2022editing}, \texttt{\textsc{FusionChat}-7B TIES}~\citep{yadav2023ties}, \texttt{\textsc{FusionChat}-7B DARE}~\citep{yu2023language}, and our \texttt{\textsc{FusionChat}-7B \textsc{VaRM}}.

\subsection{Overall Results}
\label{sec:overall_results}

In Table \ref{tab:main_results}, we present the overall results of \textsc{FusionChat} compared to baselines of different scales and categories across various domains of MT-Bench. Our observations are as follows. First, we note distinct performance among the three source LLMs across all domains, with \texttt{OpenChat-3.5-7B} exhibiting balanced performance despite its smaller scale. Second, after fine-tuning using the casual language model objective on our high-quality training dataset, the resulting model (\texttt{OpenChat-3.5-7B CLM}) achieves an increased average performance from 7.79 to 7.95, although this improvement is relatively modest and inconsistent across distinct domains. Third, in the category of fine-tuned target LLMs, \texttt{OpenChat-3.5-7B Multi} achieves a relative performance gain of 1.38\% over \texttt{OpenChat-3.5-7B CLM}. Notably, \texttt{OpenChat-3.5-7B Mixtral} and \texttt{OpenChat-3.5-7B Solar}, two target LLMs obtained by pairwise knowledge fusion, outperform \texttt{OpenChat-3.5-7B Multi}. Moreover, these target LLMs demonstrate individual strengths in different domains, providing a foundation for subsequent integration into a more powerful LLM. For instance, \texttt{OpenChat-3.5-7B Mixtral} excels in the \textit{reasoning} domain, surpassing \texttt{OpenChat-3.5-7B CLM} by an average of 12.58\%, while \texttt{OpenChat-3.5-7B Solar} achieves the highest scores in both the \textit{extraction} and \textit{STEM} domains, with 8.70\% and 9.53\% relative performance enhancements, respectively.

\begin{table*}[thbp]
\centering
\resizebox{1\linewidth}{!}{
\begin{tabular}{l|cccccccc|c}
\toprule
\textbf{Models} & \textbf{Writing} & \textbf{Roleplay} & \textbf{Reasoning} & \textbf{Math} & \textbf{Coding} & \textbf{Extraction} & \textbf{STEM} & \textbf{Humanities} & \textbf{Avg.} \\ \hline \hline
\multicolumn{10}{c}{\emph{Closed-source LLMs}} \\ \hline
GPT-4 (March) & 9.65 & 8.90 & 9.00 & 6.80 & 8.55 & 9.38 & 9.70 & 9.95 & 8.96 \\
GPT-3.5 (March) & 9.20 & 8.40 & 5.65 & 6.30 & 6.90 & 8.85 & 8.70 & 9.55 & 7.94 \\ 
Claude-1.0 & 9.50 & 8.50 & 5.95 & 4.80 & 6.25 & 8.80 & 9.70 & 9.70 & 7.90 \\ \hline
\multicolumn{10}{c}{\emph{Source LLMs}} \\ \hline
NH2-Mixtral-8x7B & 9.70 & 9.05 & 6.65 & 6.75 & 6.00 & 8.80 & 9.70 & 9.95 & 8.33 \\
NH2-Solar-10.7B & 9.50 & 8.40 & 7.35 & 5.55 & 4.00 & 7.85 & 8.80 & 9.80 & 7.66 \\
OpenChat-3.5-7B & 9.00 & 7.85 & 7.75 & 5.30 & 5.50 & 8.45 & 8.55 & 9.90 & 7.79 \\ \hline
\multicolumn{10}{c}{\emph{Fine-tuned target LLMs}} \\ \hline
OpenChat-3.5-7B CLM & 8.80 & 8.60 & 7.55 & 5.88 & 5.45 & 8.05 & 9.45 & 9.85 & 7.95 \\ \hline
\multirow{2}{*}{OpenChat-3.5-7B Multi} & 9.35 & 8.38 & 8.40 & 5.35 & 5.15 & 8.70 & 9.28 & 9.90 & 8.06 \\
 & \textcolor{blue}{(+6.25\%)} & \textcolor{red}{(-2.56\%)} & \textcolor{blue}{(+11.26\%)} & \textcolor{red}{(-9.01\%)} & \textcolor{red}{(-5.50\%)} & \textcolor{blue}{(+8.07\%)} & \textcolor{red}{(-1.80\%)} & \textcolor{blue}{(+0.51\%)} & \textcolor{blue}{(+1.38\%)} \\ \hline
\multirow{2}{*}{OpenChat-3.5-7B Mixtral} & 9.35 & 8.35 & \textbf{8.50} & 5.80 & 5.10 & 8.25 & 9.38 & 9.90 & 8.08 \\
 & \textcolor{blue}{(+6.25\%)} & \textcolor{red}{(-2.91\%)} & \textcolor{blue}{(+12.58\%)} & \textcolor{red}{(-1.36\%)} & \textcolor{red}{(-6.42\%)} & \textcolor{blue}{(+2.48\%)} & \textcolor{red}{(-0.74\%)} & \textcolor{blue}{(+0.51\%)} & \textcolor{blue}{(+1.64\%)} \\ \hline
\multirow{2}{*}{OpenChat-3.5-7B Solar} & 9.35 & 8.43 & 8.20 & 5.70 & 5.70 & \textbf{8.75} & \textbf{9.53} & 9.80 & 8.18 \\
 & \textcolor{blue}{(+6.25\%)} & \textcolor{red}{(-1.98\%)} & \textcolor{blue}{(+8.61\%)} & \textcolor{red}{(-3.06\%)} & \textcolor{blue}{(+4.59\%)} & \textcolor{blue}{(+8.70\%)} & \textcolor{blue}{(+0.85\%)} & \textcolor{red}{(-0.51\%)} & \textcolor{blue}{(+2.89\%)} \\ \hline
\multicolumn{10}{c}{\emph{Fused LLMs}} \\ \hline
\multirow{2}{*}{\textsc{FusionChat}-7B Linear} & 9.18 & 8.23 & 8.00 & 5.78 & 5.55 & 8.50 & 9.43 & 9.78 & 8.05 \\
 & \textcolor{blue}{(+4.32\%)} & \textcolor{red}{(-4.30\%)} & \textcolor{blue}{(+5.96\%)} & \textcolor{red}{(-1.70\%)} & \textcolor{blue}{(+1.83\%)} & \textcolor{blue}{(+5.59\%)} & \textcolor{red}{(-0.21\%)} & \textcolor{red}{(-0.71\%)} & \textcolor{blue}{(+1.26\%)} \\ \hline
\multirow{2}{*}{\textsc{FusionChat}-7B SLERP} & \textbf{9.45} & 8.73 & 8.05 & \textbf{6.20} & 5.30 & 8.45 & 9.43 & \textbf{9.93} & 8.19 \\
 & \textcolor{blue}{(+7.39\%)} & \textcolor{blue}{(+1.51\%)} & \textcolor{blue}{(+6.62\%)} & \textcolor{blue}{(+5.44\%)} & \textcolor{red}{(-2.75\%)} & \textcolor{blue}{(+4.97\%)} & \textcolor{red}{(-0.21\%)} & \textcolor{blue}{(+0.81\%)} & \textcolor{blue}{(+3.02\%)} \\ \hline
\multirow{2}{*}{\textsc{FusionChat}-7B TA} & 9.35 & 8.33 & 8.15 & 5.85 & 6.05 & 8.45 & 9.50 & 9.90 & 8.20 \\
 & \textcolor{blue}{(+6.25\%)} & \textcolor{red}{(-3.14\%)} & \textcolor{blue}{(+7.95\%)} & \textcolor{red}{(-0.51\%)} & \textcolor{blue}{(+11.01\%)} & \textcolor{blue}{(+4.97\%)} & \textcolor{blue}{(+0.53\%)} & \textcolor{blue}{(+0.51\%)} & \textcolor{blue}{(+3.14\%)} \\ \hline
\multirow{2}{*}{\textsc{FusionChat}-7B TIES} & 9.38 & 8.58 & 8.15 & 5.80 & 5.85 & 8.40 & 9.45 & 9.88 & 8.18 \\
 & \textcolor{blue}{(+6.59\%)} & \textcolor{red}{(-0.23\%)} & \textcolor{blue}{(+7.95\%)} & \textcolor{red}{(-1.36\%)} & \textcolor{blue}{(+7.34\%)} & \textcolor{blue}{(+4.35\%)} & \textcolor{blue}{(+0.00\%)} & \textcolor{blue}{(+0.30\%)} & \textcolor{blue}{(+2.89\%)} \\ \hline
\multirow{2}{*}{\textsc{FusionChat}-7B DARE} & \textbf{9.45} & \textbf{8.78} & 7.75 & 5.80 & 5.80 & 8.45 & 9.45 & 9.65 & 8.14 \\
 & \textcolor{blue}{(+7.39\%)} & \textcolor{blue}{(+2.09\%)} & \textcolor{blue}{(+2.65\%)} & \textcolor{red}{(-1.36\%)} & \textcolor{blue}{(+6.42\%)} & \textcolor{blue}{(+4.97\%)} & \textcolor{blue}{(+0.00\%)} & \textcolor{red}{(-2.03\%)} & \textcolor{blue}{(+2.39\%)} \\ \hline
\multirow{2}{*}{\textsc{FusionChat}-7B \textsc{VaRM}} & 9.20 & 8.63 & 8.00 & 6.15 & \textbf{6.15} & 8.50 & 9.30 & 9.85 & \textbf{8.22} \\
 & \textcolor{blue}{(+4.55\%)} & \textcolor{blue}{(+0.35\%)} & \textcolor{blue}{(+5.96\%)} & \textcolor{blue}{(+4.59\%)} & \textcolor{blue}{(+12.84\%)} & \textcolor{blue}{(+5.59\%)} & \textcolor{red}{(-1.59\%)} & \textcolor{blue}{(+0.00\%)} & \textcolor{blue}{(+3.40\%)} \\ \hline
\bottomrule
\end{tabular}
}
\caption{Overall results of the proposed \textsc{FusionChat} compared to baselines of different scales and categories across various domains of MT-Bench. Percentages indicate the rate of improvement (in blue)/decrease (in red) compared to \texttt{OpenChat-3.5-7B CLM}.}
\label{tab:main_results}
\vspace{0.1cm}
\end{table*}

The final fused LLM \texttt{\textsc{FusionChat}-7B} is obtained by merging \texttt{OpenChat-3.5-7B Mixtral} and \texttt{OpenChat-3.5-7B Solar} in the parameter space, where various merging methods are explored.
It is observed that \texttt{\textsc{FusionChat}-7B} with \texttt{SLERP}, \texttt{TA}, and our \texttt{\textsc{VaRM}} outperform all the fine-tuned target LLMs, showcasing \textsc{FusionChat}'s ability to integrate the unique strengths and collective capabilities of different target LLMs.
In contrast, merging methods such as \texttt{Linear} and \texttt{DARE} tend to result in degraded performance. Since the target LLMs exhibit varying parameter variations, designing fine-grained merging weights is crucial for effectively combining their respective advantages. 
Therefore, methods like \texttt{Linear}, which involves manual weight assignment, and \texttt{DARE}, which eliminates a subset of model parameters before merging, are deemed inappropriate for \textsc{FusionChat}.

We further demonstrate that \texttt{\textsc{FusionChat}-7B} with \texttt{\textsc{VaRM}} consistently outperforms all other merging methods, achieving an average evaluation score of 8.22. This score not only surpasses \texttt{GPT-3.5 (March)}'s score of 7.94, but also approaches the score of the current state-of-the-art (SOTA) open-source chat LLM, \texttt{NH2-Mixtral-8X7B}, which stands at 8.33. This confirms the effectiveness of the proposed \textsc{VaRM} method in utilizing the variation ratio of each parameter matrix to allocate different merging weights, thereby blending updated knowledge at a fine-grained matrix level.

\begin{table*}[thbp]
\centering
\resizebox{0.99\linewidth}{!}{
\begin{tabular}{l|cccccccc|c}
\toprule
\textbf{Granularity} & \textbf{Writing} & \textbf{Roleplay} & \textbf{Reasoning} & \textbf{Math} & \textbf{Coding} & \textbf{Extraction} & \textbf{STEM} & \textbf{Humanities} & \textbf{Avg.} \\ \hline \hline
Model & 8.93 & 8.58 & \textbf{8.20} & 5.73 & 5.45 & \textbf{8.55} & \textbf{9.45} & \textbf{9.90} & 8.10 \\
Layer & \textbf{9.43} & 8.73 & 8.00 & \textbf{6.15} & 5.25 & \textbf{8.55} & 9.18 & \textbf{9.90} & 8.15 \\
Matrix & 9.20 & 8.63 & 8.00 & \textbf{6.15} & \textbf{6.15} & 8.50 & 9.30 & 9.85 & \textbf{8.22} \\
Parameter & 1.00 & 1.00 & 1.00 & 1.00 & 1.00 & 1.00 & 1.00 & 1.00 & 1.00 \\ \hline
\bottomrule
\end{tabular}
}
\caption{Results of \texttt{\textsc{FusionChat}-7B \textsc{VaRM}} with \textsc{VaRM} adopting different merging granularities of parameter units across various domains of MT-Bench.}
\label{tab:merge_granularity_cat_results}
\vspace{-0.3cm}
\end{table*}

\subsection{Merging Granularities in \textsc{VaRM}}
\label{sec:analysis_of_merging_granularity}

Since the merging granularity of the parameter unit $\theta_{j,m}$ in Eq. \ref{eqn:varn} can be adaptively adjusted, we investigate its influence on the final performance of \texttt{\textsc{FusionChat-7B} \textsc{VaRM}}. 

\begin{wrapfigure}{r}{0.45\textwidth}
    \centering
    \includegraphics[width=0.95\linewidth]{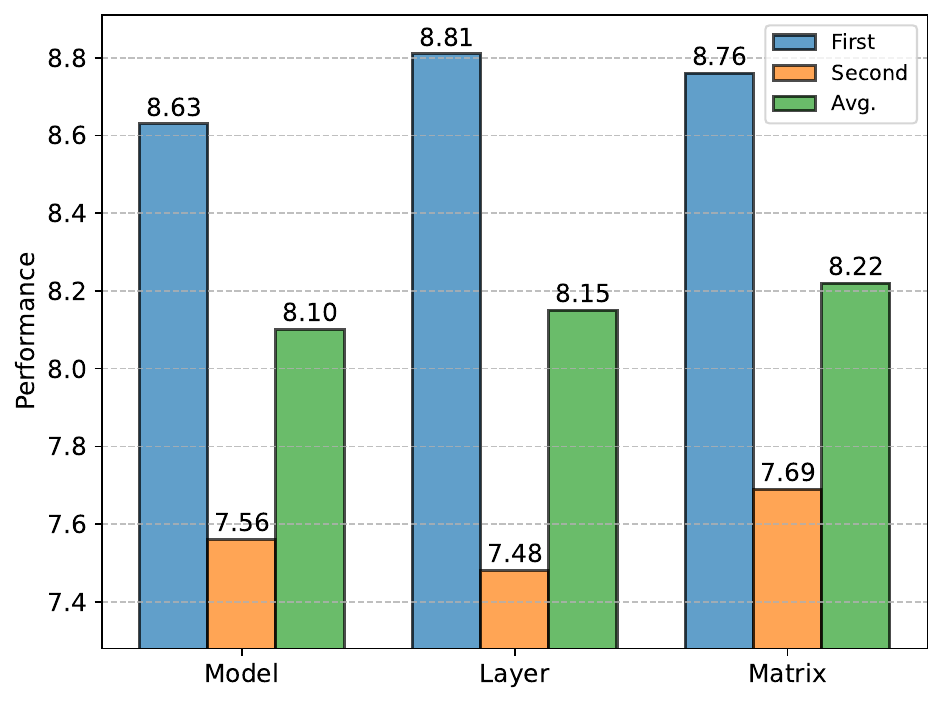}
    \caption{Performance of \texttt{\textsc{FusionChat}-7B \textsc{VaRM}} by using varying merging granularities of parameter groups on different dialogue turns in MT-Bench.}
    \label{fig:turn_analysis}
    \vspace{-0.3cm}
\end{wrapfigure}

In Table \ref{tab:merge_granularity_cat_results}, we observe a consistent improvement in average performance when transitioning the granularity of merging weights from model level to layer level, and then to matrix level. This suggests that the assignment of fine-grained merging weights is effective for integrating knowledge from multiple target LLMs. 
However, when the granularity is reduced to the parameter level, we observe a notable decline in performance. This may be attributed to extreme merging weights assigned to specific parameters, which disrupts correlations among other parameters.

We further investigate the impact of varying merging granularities on the performance of different dialogue turns. Figure \ref{fig:turn_analysis} illustrates that as the merging granularity progressively decreases from the model level to the layer level and then to the matrix level, the performance of the first turn first experiences enhancement and then declines, while the performance of the second turn exhibits the opposite trend. Despite this fluctuation, there is a consistent improvement in overall performance. These findings suggest that \textsc{VaRM} at the matrix granularity adeptly captures the complex dynamics among multiple dialogue turns, thereby leading to optimal overall performance.

\section{Conclusion}
In this work, we propose an extended framework of \textsc{FuseLLM} to integrate the collective knowledge and individual strengths of multiple structure and scale-varied chat LLMs into a more powerful chat LLM, resulting in \textsc{FusionChat}. \textsc{FusionChat} adopts a fuse-then-merge strategy with two main stages. Firstly, it undertakes pairwise knowledge fusion for source LLMs to derive multiple target LLMs of identical structure and size via lightweight fine-tuning. Then, these target LLMs are merged within the parameter space, wherein we propose a novel method \textsc{VaRM} for determining the merging weights based on the variation ratio of parameter matrices before and after fine-tuning. Experimental results spanning various chat domains demonstrate the superiority of \textsc{FusionChat} across different model scales, even surpassing \texttt{GPT-3.5 (March)} and approaching \texttt{Mixtral-8x7B-Instruct}.

Moreover, we argue that the concept of knowledge fusion adopted by both \textsc{FusionChat} and \textsc{FuseLLM} shares a fundamentally similar purpose with other related topics, such as the recently popular topic of mixture of experts (MoEs), because they all aim to leverage the strengths of multiple models (experts). However, while MoEs require loading multiple experts during inference, which has higher memory requirements, knowledge fusion supports the integration of multiple LLMs with diverse architectures into a single LLM without any additional memory requirement, making it more memory-efficient. To the best of our knowledge, MoEs typically employ more than six experts, while \textsc{FusionChat} and \textsc{FuseLLM} only fuse three source LLMs. In future work, we will further explore fusing more source LLMs to fully harness the potential of knowledge fusion for LLMs.

\bibliographystyle{apalike}
\bibliography{custom}

\appendix

\section{Details of Training Dataset}
\label{appendix: details_of_training_dataset}

We curated a comprehensive training dataset, \textsc{FusionChat-Mixture}, from various sources. This dataset covers different styles and capabilities, featuring both human-written and model-generated, and spanning general instruction-following and specific skills. These sources include: 
\begin{itemize}[topsep=0pt,labelindent=0.5cm,leftmargin=*,wide=0pt]
\item \textbf{Orca-Best}\footnote{\url{https://huggingface.co/datasets/shahules786/orca-best}}: We sampled 20,000 examples from Orca-Best, which is filtered from the original GPT-4 (1M) partition of Orca~\citep{mukherjee2023orca} based on maximum length and embedding clustering of instructions.

\item \textbf{Capybara}\footnote{\url{https://huggingface.co/datasets/LDJnr/Capybara}}: We incorporated all the 16,000 examples of Capybara, which is a high-quality collection of multi-turn synthetic conversations.

\item \textbf{No-Robots}\footnote{\url{https://huggingface.co/datasets/HuggingFaceH4/no_robots}}: We included all the 9,500 examples of No-Robots, which is a dataset created by skilled human annotators for supervised fine-tuning. 

\item \textbf{ShareGPT-GPT4}\footnote{\url{https://huggingface.co/datasets/shibing624/sharegpt_gpt4}}: We utilized all 6,200 examples from ShareGPT-GPT4, which exclusively uses dialogues generated by GPT-4 in ShareGPT.

\item \textbf{Oasst-Top1}\footnote{\url{https://huggingface.co/datasets/OpenAssistant/oasst_top1_2023-08-25}}: We selected 5,000 examples from Oasst-Top1, which is a refined version of Oasst1~\citep{kopf2023openassistant}, a human-annotated assistant-style conversation dataset.

\item \textbf{MetaMathQA}~\footnote{\url{https://huggingface.co/datasets/meta-math/MetaMathQA}}: We sampled 10,000 examples from MetaMathQA~\citep{yu2023metamath}, which is augmented from the GSM8K~\citep{cobbe2021training} and MATH~\citep{hendrycks2021measuring} datasets for mathematics problem-solving.

\item \textbf{OSS-Instruct}~\footnote{\url{https://huggingface.co/datasets/ise-uiuc/Magicoder-OSS-Instruct-75K}}: We chose 10,000 examples from OSS-Instruct~\citep{wei2023magicoder}, which contains code instruction data synthesized from open-source code snippets.

\item \textbf{Evol-Alpaca}~\footnote{\url{https://huggingface.co/datasets/theblackcat102/evol-codealpaca-v1}}: We sampled 10,000 examples from Evol-Alpaca, which is a code instruction dataset generated by GPT-4 with evol-instruct proposed by WizardCoder~\citep{luo2023wizardcoder}.

\item \textbf{Python-Code}~\footnote{\url{https://huggingface.co/datasets/ajibawa-2023/Python-Code-23k-ShareGPT}}: We selected 10,000 examples from Python-Code, which comprises instructions and responses generated by GPT-3.5 and GPT-4 for python code generation.

\end{itemize}

We followed the data processing code in Vicuna~\citep{vicuna2023} to clean instances containing non-English or special characters. Then, we split long conversations into blocks with a maximum length of 2048 tokens, resulting in the final \textsc{FusionChat Mixture} with 95,000 examples.

\section{Case Studies}
\label{appendix: case_studies}

We present case studies to demonstrate the individual strengths of target LLMs (\texttt{OpenChat-3.5-7B Mixtral} and \texttt{OpenChat-3.5-7B Solar}) obtained from knowledge fusion of source LLMs, and show the collective knowledge and strengths of \textsc{FusionChat} (\texttt{\textsc{FusionChat}-7B \textsc{VaRM}}) obtained by further merging target LLMs. \texttt{OpenChat-3.5-7B CLM} is used as the baseline for comparison.

\begin{table*}[thbp]
\centering
\renewcommand\arraystretch{1.1}
\resizebox{0.95\linewidth}{!}{
\begin{tabular}{p{15cm}}
\hline
\textbf{OpenChat-3.5-7B CLM; Roleplay; ID 93;}  \\ \hline
\textbf{Turn 1 Question:} Imagine yourself as a doctor tasked with devising innovative remedies for various ailments and maladies. Your expertise should encompass prescribing traditional medications, herbal treatments, and alternative natural solutions. Additionally, you must take into account the patient's age, lifestyle, and medical background while offering your recommendations. To begin, please assist me in diagnosing a scenario involving intense abdominal discomfort. \\ \hline
\textbf{Turn 1 Answer:} As a doctor, I would first gather information about the patient's symptoms, medical history, and lifestyle. In the case of intense abdominal discomfort, I would consider several potential causes, such as indigestion, gas, food poisoning, or an inflammatory bowel disease. I would then recommend a combination of traditional medications, herbal treatments, and alternative natural solutions, taking into account the patient's age, lifestyle, and medical background.
For a patient with intense abdominal discomfort due to indigestion or gas, I would recommend the following remedies:
1. Traditional medications: Over-the-counter antacids such as Maalox or Mylanta can help alleviate indigestion and gas.
2. Herbal treatments: Chamomile tea or peppermint tea can help soothe the stomach and reduce abdominal discomfort.
3. Alternative natural solutions: Encourage the patient to avoid trigger foods, such as spicy or fatty foods, and to consume smaller meals more frequently. Additionally, they can try gentle exercises like walking or yoga to improve digestion.
For a patient with food poisoning, I would recommend the following remedies:
1. Traditional medications: Depending on the severity of the symptoms, the patient may need prescription antibiotics. Over-the-counter medications such as loperamide can help control diarrhea.
2. Herbal treatments: Ginger tea or lemon water can help soothe the stomach and reduce nausea.
3. Alternative natural solutions: Encourage the patient to stay hydrated by consuming clear fluids like water and electrolyte drinks. Once symptoms subside, probiotics can be taken to restore gut flora balance.
For a patient with an inflammatory bowel disease, I would recommend the following remedies:
1. Traditional medications: Prescription medications such as aminosalicylates or corticosteroids may be necessary to reduce inflammation and control symptoms.
2. Herbal treatments: Some herbs, such as turmeric and slippery elm, have anti-inflammatory properties that may help alleviate symptoms. However, it is essential to consult with a healthcare professional before using these herbs, as they can interact with other medications or have side effects.
3. Alternative natural solutions: Encourage the patient to maintain a healthy diet, rich in fiber, and low in processed foods. Stress management techniques, such as meditation or yoga, can also help manage symptoms.
In all cases, it is essential to monitor the patient's progress and adjust the treatment plan as needed. If symptoms worsen or do not improve within a reasonable timeframe, further testing and specialist referral may be necessary. \\ \hline
\textbf{Turn 1 Judgment:} The assistant's response is highly detailed, accurate, and relevant to the user's question. It provides a comprehensive approach to diagnosing and treating intense abdominal discomfort, considering various potential causes. The assistant also takes into account the need for traditional medications, herbal treatments, and alternative natural solutions, as requested by the user. The assistant's response is also creative in its approach to providing a variety of treatment options for each potential cause of abdominal discomfort. The assistant's response is also helpful in that it provides clear and actionable advice for each potential cause of the discomfort. Therefore, the assistant's response is of high quality. \textcolor{blue}{Rating: [[9]]} \\ \hline
\textbf{Turn 2 Question:} But I have been pregnant for 20 weeks and I am allergic to many medicines. \\ \hline
\textbf{Turn 2 Answer:} I am not a doctor, but I can try to provide some general guidance. If you are pregnant and experiencing abdominal discomfort, it's essential to consult with your healthcare provider before starting any treatment. They will consider your pregnancy and medication allergies when recommending a course of action.
In the meantime, here are some general suggestions for managing abdominal discomfort during pregnancy:
1. Eat smaller meals throughout the day to avoid overfilling your stomach.
2. Avoid spicy, fatty, or greasy foods that may worsen indigestion.
3. Stay hydrated by drinking plenty of water.
4. Practice gentle exercises, such as walking or prenatal yoga, to aid digestion.
However, it's crucial to consult with your healthcare provider for personalized advice tailored to your specific situation. \\ \hline
\textbf{Turn2 Judgment:} The assistant's response is relevant, accurate, and detailed. It provides general advice for managing abdominal discomfort during pregnancy, while emphasizing the importance of consulting with a healthcare provider due to the user's specific circumstances (pregnancy and allergies to many medicines). The assistant also acknowledges its limitations, stating that it is not a doctor and can only provide general guidance. The response is helpful and takes into account the user's specific situation, but it could have been more creative by suggesting more specific natural remedies that are safe during pregnancy. Overall, the assistant's response is of high quality. \textcolor{blue}{Rating: [[8]]} \\ \hline
\end{tabular}
}
\caption{Case studies on roleplay. The answers are generated by \texttt{OpenChat-3.5-7B CLM}.}
\label{tab:case-1-clm}
\end{table*}

\begin{table*}[thbp]
\centering
\renewcommand\arraystretch{1.1}
\resizebox{0.95\linewidth}{!}{
\begin{tabular}{p{15cm}}
\hline
\textbf{OpenChat-3.5-7B Mixtral; Roleplay; ID 93;}  \\ \hline
\textbf{Turn 1 Question:} Imagine yourself as a doctor tasked with devising innovative remedies for various ailments and maladies. Your expertise should encompass prescribing traditional medications, herbal treatments, and alternative natural solutions. Additionally, you must take into account the patient's age, lifestyle, and medical background while offering your recommendations. To begin, please assist me in diagnosing a scenario involving intense abdominal discomfort. \\ \hline
\textbf{Turn 1 Answer:} As a doctor, I would first gather information about the patient's medical history, the duration and severity of the abdominal discomfort, and any other symptoms they might be experiencing. After obtaining a thorough understanding of the patient's situation, I would consider various factors, such as age, lifestyle, and medical background, to determine the most appropriate treatment plan.
1. Traditional Medications:
If the abdominal discomfort is due to a condition such as indigestion or irritable bowel syndrome, over-the-counter medications like antacids or anti-spasmodics may be prescribed to alleviate symptoms. For more severe cases, prescription medications may be necessary, depending on the underlying cause. It is essential to monitor the patient's response to the medication and adjust the dosage or treatment plan accordingly.
2. Herbal Treatments:
Herbal remedies can be a valuable addition to a treatment plan for abdominal discomfort. Chamomile, for example, has anti-inflammatory and calming properties that can help soothe the digestive system. Peppermint oil may also be beneficial for relieving abdominal pain and reducing bloating. However, patients should consult with a healthcare professional before incorporating herbal treatments into their regimen, as some herbs may interact with other medications or have contraindications for certain individuals.
3. Alternative Natural Solutions:
In addition to herbal remedies, there are other natural approaches to managing abdominal discomfort. Probiotics, for instance, can help restore balance to the gut microbiome, which may alleviate symptoms in cases of indigestion or irritable bowel syndrome. Dietary modifications, such as reducing intake of high-fat, high-sugar, or processed foods, and increasing consumption of fiber-rich foods, can also help improve digestive health.
4. Lifestyle Recommendations:
Encouraging the patient to adopt a healthier lifestyle may also contribute to alleviating abdominal discomfort. Regular exercise, stress reduction techniques (such as yoga or meditation), and maintaining a healthy weight can all support optimal digestive function.
In summary, the appropriate treatment for abdominal discomfort depends on the underlying cause and the patient's unique circumstances. A combination of traditional medications, herbal treatments, alternative natural solutions, and lifestyle adjustments may be recommended to effectively manage the condition. As a doctor, I would continually assess the patient's response to the treatment plan and make any necessary adjustments to ensure their comfort and well-being. \\ \hline
\textbf{Turn 1 Judgment:} The assistant's response is comprehensive, accurate, and highly relevant to the user's question. It provides a detailed approach to diagnosing and treating abdominal discomfort, considering the patient's age, lifestyle, and medical background. The assistant suggests traditional medications, herbal treatments, alternative natural solutions, and lifestyle adjustments, which shows a deep understanding of the topic. The assistant also emphasizes the importance of monitoring the patient's response to the treatment plan, which is a crucial aspect of patient care. The response is creative in its approach to combining different treatment methods and is detailed in its explanation of each method. Therefore, the assistant's response is of high quality. \textcolor{blue}{Rating: [[10]]} \\ \hline
\textbf{Turn 2 Question:} But I have been pregnant for 20 weeks and I am allergic to many medicines. \\ \hline
\textbf{Turn 2 Answer:} I'm not a doctor, but I can try to provide you with some general guidance based on the information you've provided. Pregnancy can indeed present unique challenges when it comes to managing pain or discomfort, especially if you have allergies to certain medications.
1. Consideration of Pregnancy:
During pregnancy, it's crucial to take into account the safety of treatments and medications for both the mother and the developing baby. Always consult with your healthcare provider before starting any new treatment, as they will be able to recommend options that are safe for you and your baby.
2. Allergy Considerations:
If you are allergic to many medications, it's important to be cautious when trying new treatments. Make sure to inform your healthcare provider about your allergies, as they can help identify alternative options that are less likely to cause adverse reactions.
3. Herbal Remedies and Natural Solutions:
As mentioned earlier, herbal remedies and natural solutions can be a valuable addition to a treatment plan during pregnancy. However, it's essential to consult with your healthcare provider before using any herbal products, as some can be harmful during pregnancy.
4. Alternative Treatments:
In cases where traditional medications and herbal remedies are not suitable, alternative treatments such as acupuncture, massage, or physical therapy may be considered, provided they are approved by your healthcare provider.
5. Communication with Your Healthcare Provider:
It's vital to maintain open and honest communication with your healthcare provider throughout your pregnancy. Discuss your symptoms, concerns, and any allergies you may have, so they can tailor a treatment plan that addresses your needs while ensuring the safety of both you and your baby.
Remember, it's important to consult with a qualified healthcare professional before making any decisions about your treatment plan during pregnancy. They will be able to provide personalized guidance based on your specific circumstances and medical history. \\ \hline
\textbf{Turn2 Judgment:} The assistant's response is highly relevant, accurate, and detailed. It provides a comprehensive answer to the user's question, taking into account the user's pregnancy and allergies to medications. The assistant emphasizes the importance of consulting with a healthcare provider before starting any new treatment, which is crucial advice for a pregnant woman. The assistant also suggests alternative treatments such as acupuncture, massage, or physical therapy, which could be helpful for the user. The assistant's response is also creative, as it suggests a variety of potential solutions for the user's problem. Overall, the assistant's response is very helpful and informative. \textcolor{blue}{Rating: [[9]]} \\ \hline
\end{tabular}
}
\caption{Case studies on roleplay. The answers are generated by \texttt{OpenChat-3.5-7B Mixtral}.}
\label{tab:case-1-mix}
\end{table*}

\begin{table*}[thbp]
\centering
\renewcommand\arraystretch{1.1}
\resizebox{0.95\linewidth}{!}{
\begin{tabular}{p{15cm}}
\hline
\textbf{OpenChat-3.5-7B Solar; Roleplay; ID 93;}  \\ \hline
\textbf{Turn 1 Question:} Imagine yourself as a doctor tasked with devising innovative remedies for various ailments and maladies. Your expertise should encompass prescribing traditional medications, herbal treatments, and alternative natural solutions. Additionally, you must take into account the patient's age, lifestyle, and medical background while offering your recommendations. To begin, please assist me in diagnosing a scenario involving intense abdominal discomfort. \\ \hline
\textbf{Turn 1 Answer:} As a doctor, I would first gather information about the patient's medical history, the duration and severity of the abdominal discomfort, and any other symptoms they may be experiencing. After obtaining a thorough understanding of the patient's situation, I would consider various factors, such as age, lifestyle, and medical background, to determine the most appropriate treatment plan.
1. Traditional Medications:
If the abdominal discomfort is due to a condition such as indigestion or irritable bowel syndrome, over-the-counter medications like antacids or anti-spasmodics may be prescribed to alleviate symptoms.
2. Herbal Treatments:
Chamomile tea is a well-known herbal remedy for digestive issues. It has anti-inflammatory and antispasmodic properties that can help soothe the gut and alleviate abdominal discomfort. Peppermint oil can also be effective in reducing symptoms of irritable bowel syndrome.
3. Alternative Natural Solutions:
Probiotics can help restore balance to the gut microbiome, which may alleviate abdominal discomfort. Additionally, consuming ginger, which has anti-inflammatory and antioxidant properties, may help reduce symptoms of indigestion and nausea.
Considering the patient's age, lifestyle, and medical background is crucial for determining the best course of action. For example, older individuals or those with severe abdominal pain may require additional testing or prescription medications. Those with a history of certain medical conditions may need to avoid specific herbal treatments or natural solutions.
In conclusion, a multifaceted approach that combines traditional medications, herbal treatments, and alternative natural solutions, tailored to the patient's unique needs, can be an effective way to address abdominal discomfort. However, it is essential to consult with a healthcare professional before starting any new treatment plan. \\ \hline
\textbf{Turn 1 Judgment:} The assistant's response is comprehensive, accurate, and relevant to the user's question. It provides a detailed approach to diagnosing and treating abdominal discomfort, considering the patient's age, lifestyle, and medical background. The assistant suggests traditional medications, herbal treatments, and alternative natural solutions, explaining how each could be beneficial. The assistant also emphasizes the importance of consulting with a healthcare professional before starting any new treatment plan, which is a crucial point. The response is well-structured and informative, demonstrating a good understanding of the topic. Therefore, I would rate this response highly. \textcolor{blue}{Rating: [[9]]} \\ \hline
\textbf{Turn 2 Question:} But I have been pregnant for 20 weeks and I am allergic to many medicines. \\ \hline
\textbf{Turn 2 Answer:} I'm not a doctor, but I can try to provide some general guidance. It's important to consult with your healthcare provider when dealing with pregnancy-related symptoms, as some treatments may not be safe for the mother or fetus.
1. Traditional Medications:
During pregnancy, certain medications are considered safe, while others are not. Be sure to inform your healthcare provider about your allergies and any medications you've taken during pregnancy. They will be able to recommend safe options based on your specific situation.
2. Herbal Treatments:
Some herbal remedies may be safe to use during pregnancy, while others should be avoided. Ginger, for example, is often used to treat nausea and morning sickness and is generally considered safe in moderate amounts. However, it's important to consult with your healthcare provider before using any herbal treatments during pregnancy.
3. Alternative Natural Solutions:
Pregnant women should be cautious with alternative natural solutions, as some may not be safe during pregnancy. For example, certain essential oils should not be used during pregnancy. Always consult with your healthcare provider before trying any new treatments.
In summary, it's essential to consult with your healthcare provider when dealing with pregnancy-related symptoms, especially if you have allergies to medications or other substances. They will be able to recommend safe and effective treatment options tailored to your specific needs. \\ \hline
\textbf{Turn2 Judgment:} The assistant's response is relevant, accurate, and detailed. It provides a comprehensive answer to the user's question, taking into account the user's pregnancy and allergies. The assistant correctly advises the user to consult with a healthcare provider before starting any new treatment, which is crucial given the user's condition. The assistant also provides examples of treatments that might be safe during pregnancy, but again emphasizes the importance of consulting with a healthcare provider. The assistant's response is also creative, as it provides a variety of potential solutions, including traditional medications, herbal treatments, and alternative natural solutions. \textcolor{blue}{Rating: [[9]]} \\ \hline
\end{tabular}
}
\caption{Case studies on roleplay. The answers are generated by \texttt{OpenChat-3.5-7B Solar}.}
\label{tab:case-1-so}
\end{table*}

\begin{table*}[thbp]
\centering
\renewcommand\arraystretch{1.1}
\resizebox{0.95\linewidth}{!}{
\begin{tabular}{p{15cm}}
\hline
\textbf{\textsc{FusionChat}-7B \textsc{VaRM}; Roleplay; ID 93;}  \\ \hline
\textbf{Turn 1 Question:} Imagine yourself as a doctor tasked with devising innovative remedies for various ailments and maladies. Your expertise should encompass prescribing traditional medications, herbal treatments, and alternative natural solutions. Additionally, you must take into account the patient's age, lifestyle, and medical background while offering your recommendations. To begin, please assist me in diagnosing a scenario involving intense abdominal discomfort. \\ \hline
\textbf{Turn 1 Answer:} As a doctor, I would first gather information about the patient's medical history, the duration and severity of the abdominal discomfort, and any other symptoms they may be experiencing. After obtaining a thorough understanding of the patient's situation, I would consider various factors, such as age, lifestyle, and medical background, to determine the most appropriate treatment plan.
1. Traditional Medications:
If the abdominal discomfort is due to a condition such as indigestion or irritable bowel syndrome, over-the-counter medications like antacids or anti-spasmodics may be prescribed to alleviate symptoms. For more severe cases, prescription medications may be necessary, depending on the underlying cause. It is essential to monitor the patient's response to the medication and adjust the dosage or treatment plan accordingly.
2. Herbal Treatments:
Herbal remedies can be a valuable addition to a treatment plan for abdominal discomfort. Chamomile, for example, has anti-inflammatory and calming properties that can help soothe the digestive system. Peppermint oil may also be beneficial for relieving abdominal pain and reducing bloating. However, patients should consult with a healthcare professional before incorporating herbal treatments into their regimen, as some herbs may interact with other medications or have contraindications for certain individuals.
3. Alternative Natural Solutions:
In addition to herbal remedies, there are other natural solutions that can help alleviate abdominal discomfort. These may include dietary modifications, such as avoiding trigger foods, increasing water intake, and consuming more fiber. Probiotics, which are beneficial bacteria found in fermented foods like yogurt, can also support gut health and alleviate symptoms. Stress management techniques, such as deep breathing exercises or yoga, may also help reduce abdominal discomfort in some cases.
In conclusion, the treatment plan for intense abdominal discomfort will depend on the underlying cause and the patient's individual circumstances. A combination of traditional medications, herbal treatments, and alternative natural solutions may be recommended to provide the most effective relief. Regular follow-ups with the patient will be crucial to assess their progress and make any necessary adjustments to their treatment plan. \\ \hline
\textbf{Turn 1 Judgment:} The assistant's response is comprehensive, accurate, and highly relevant to the user's question. It provides a detailed approach to diagnosing and treating abdominal discomfort, considering the patient's medical history, age, and lifestyle. The assistant also offers a variety of treatment options, including traditional medications, herbal treatments, and alternative natural solutions, explaining each in depth. The assistant emphasizes the importance of consulting with a healthcare professional before starting any treatment, which is a crucial point. The response is well-structured and informative, providing a balanced view of different treatment options. \textcolor{blue}{Rating: [[10]]} \\ \hline
\textbf{Turn 2 Question:} But I have been pregnant for 20 weeks and I am allergic to many medicines. \\ \hline
\textbf{Turn 2 Answer:} I'm not a doctor, but I can try to help you with some general suggestions. If you're pregnant and experiencing abdominal discomfort, it's crucial to consult with your healthcare provider before trying any remedies. They will be able to take into account your specific situation, including your allergies and pregnancy status, and recommend the safest and most effective treatment options.
1. Dietary modifications:
Pregnant women should avoid trigger foods that may cause abdominal discomfort, such as spicy or fatty foods, beans, and carbonated drinks. Eating smaller meals more frequently, avoiding lying down after eating, and chewing food thoroughly may also help.
2. Hydration:
Drinking plenty of water is essential during pregnancy, as it can help prevent dehydration, which may contribute to abdominal discomfort. Aim for at least eight 8-ounce glasses of water per day.
3. Gentle exercise:
Light exercises like walking or prenatal yoga can help improve blood circulation and alleviate abdominal discomfort. However, it is essential to consult with your healthcare provider before starting any exercise routine during pregnancy.
4. Heat or cold therapy:
Applying a heating pad or warm compress to your abdomen may help relieve muscle cramps or discomfort. Alternatively, a cold compress can help reduce inflammation and soothe the area. Ensure the temperature is not too hot, and do not apply it directly to the skin – use a towel or cloth as a barrier.
5. Herbal remedies:
Some herbal remedies, such as ginger, may be safe during pregnancy and can help alleviate nausea, indigestion, and abdominal discomfort. However, it is essential to consult with your healthcare provider before using any herbal treatments.
6. Probiotics:
Probiotics may help improve digestion and reduce abdominal discomfort during pregnancy. Look for prenatal probiotics specifically designed for pregnant women.
Remember that it is essential to consult with your healthcare provider before trying any remedies, as they can provide personalized advice based on your specific situation and pregnancy stage. \\ \hline
\textbf{Turn2 Judgment:} The assistant's response is very helpful, relevant, and detailed. It provides a comprehensive list of potential remedies for the user's abdominal discomfort during pregnancy, taking into account the user's allergies to many medicines. The assistant also emphasizes the importance of consulting with a healthcare provider before trying any remedies, which is crucial advice for a pregnant woman. The assistant's response is also creative, as it suggests a variety of solutions, including dietary modifications, hydration, gentle exercise, heat or cold therapy, herbal remedies, and probiotics. The assistant's response is accurate and shows a deep understanding of the user's situation. \textcolor{blue}{Rating: [[10]]} \\ \hline
\end{tabular}
}
\caption{Case studies on roleplay. The answers are generated by \texttt{\textsc{FusionChat}-7B \textsc{VaRM}}.}
\label{tab:case-1-fuse}
\end{table*}

\end{document}